\documentclass{article}
\usepackage[dblblindworkshop,final]{{neurips_2026}}
\workshoptitle{{SLMs for Agentic Systems, Paris, France, 2026}}
\usepackage[utf8]{inputenc}
\usepackage[T1]{fontenc}
\usepackage{hyperref}
\usepackage{url}
\usepackage{graphicx}
\usepackage{booktabs}
\usepackage{multirow}
\usepackage{amsmath}
\usepackage{amsfonts}
\usepackage{nicefrac}
\usepackage{microtype}
\usepackage{xcolor}
\usepackage[most]{tcolorbox}

\definecolor{promptframe}{HTML}{2F4858}
\definecolor{promptback}{HTML}{F7F9FA}
\definecolor{prompttitle}{HTML}{2F4858}
\definecolor{sysrole}{HTML}{8A4B08}
\definecolor{usrrole}{HTML}{1F5673}

\newtcolorbox{promptbox}[1]{
  enhanced,
  breakable,
  colback=promptback,
  colframe=promptframe,
  boxrule=0.6pt,
  arc=2pt,
  left=6pt,
  right=6pt,
  top=5pt,
  bottom=5pt,
  fonttitle=\bfseries\small,
  coltitle=white,
  title={#1},
  attach boxed title to top left={xshift=6pt,yshift=-2.5pt},
  boxed title style={
    colback=prompttitle,
    colframe=prompttitle,
    arc=1.5pt,
    left=4pt,
    right=4pt,
    top=1.5pt,
    bottom=1.5pt
  },
  fontupper=\ttfamily\scriptsize
}

\title{Verification Turns Capability Gaps into Routing Decisions: Small Language Models for Clinical Documentation}

\author{%
  Nina Fatehi \\
  Optum\\
  \texttt{nina\_fatehi@optum.com} \\
  \And
   Meysam Ghaffari \\
   Optum\\
  \texttt{seyed\_ghaffaridehkordi@optum.com} \\
  \AND
  Animesh Agarwal \\
  Optum \\
  \texttt{animesh.agarwal@optum.com} \\
   \And
  Anushka Swarup \\
   Optum\\
  \texttt{anushka\_swarup@optum.com} \\
  \And
  Carlos Morato\\
  Optum \\
  \texttt{carlos.morato@optum.com} \\
}

\begin{document}

\maketitle

\begin{abstract}
Clinical note generation from doctor--patient dialogue is high-volume and hallucination-intolerant: a
fabricated laboratory value can create patient-safety risk, not a quality regression. We ask whether an
agentic \emph{verification} architecture, rather than generator scale, can make a small language model
viable here. Holding a four-evaluator verification stack and refinement policy fixed and swapping only
the generator between a 4B-parameter model with a QLoRA adapter and a frontier baseline, the small
model reaches $0.984$ claim-level factual entailment against $0.992$ and comes within $3.3$ points on
completeness, generating entirely on-device. Its deficit is confined to \emph{inter-section structural
fidelity}---arrangement rather than clinical comprehension, and the axis on which it also falls
furthest below the corpus's reference notes. Because that deficit is an explicit verifier decision
rather than a silent quality difference, it becomes a routing policy: run the small model by default,
escalate what the verifier rejects. Measured end to end, that it achieved a similiar verified-rate acceptance to the frontier-only acceptance in this test set ($99.6\%$ vs.\ $98.4\%$ over $250$ encounters) while routing $62$, documenting three quarters without
a frontier \emph{generation} call. Ablating the loop separates architecture from scale: unrefined, the
small model is accepted on $26.8\%$ of encounters and the frontier model on $74.8\%$, so verification
does the most work where the generator is weakest. We also contribute section-scoped refinement, which
preserves unflagged sections by construction rather than instruction, and find the baseline is not
uniformly stronger---its provider refused two trauma-related transcripts the on-device model handled
and certified, one facet of a provider dependency a served adapter does not carry.
\end{abstract}

\section{Introduction}
\label{sec:intro}

Ambient clinical documentation is high-volume and hallucination-intolerant: a fabricated laboratory
value or medication can create patient-safety risk, not a quality regression, and routing every encounter
to a third-party API makes availability and data locality operational concerns rather than
afterthoughts. A recent position argues small language models (SLMs)
are the natural substrate for agentic systems on repetitive, specialized
tasks~\cite{belcak2025slmagentic}---specifically, that verification and handoff can let a small
model own most of a safety-critical workload beyond its unaided reach. We test that claim
empirically in a setting where the cost of being wrong is unusually concrete. Concretely, we ask
whether a 4B-parameter model with a QLoRA
adapter~\cite{hu2021lora,dettmers2023qlora}, embedded in an architecture that \emph{verifies} every
candidate note against explicit rubrics and revises it until it passes or is rejected, can produce
notes comparable to a frontier model on the axis that governs clinical safety: factual grounding in
the source encounter.

Our setting permits an unusually clean test. The verification stack and refinement policy are
generator-agnostic and implemented once, shared verbatim, so we swap only the generator between the
SLM and a frontier baseline (GPT-4.1), holding every evaluator, rubric, threshold, and refinement
mechanism fixed. The SLM lands within $0.008$ of the frontier model on factual
entailment and level with the reference corpus there; on completeness it trails by $3.3$ points while
exceeding the corpus by $12.7$; structural fidelity is where it falls furthest below both, and we argue
it is also the most tractable. Because the shortfall surfaces as an explicit
verifier decision rather than a silent quality difference, it converts into a deployment policy
rather than a disqualification: run the SLM by default and escalate only the notes it fails. Measured
end to end, that policy matches frontier-only quality ($99.6\%$ vs.\ $98.4\%$ acceptance) while
sending only $62$ of $250$ encounters to the frontier model, eliminating the frontier
\emph{generation} dependency for the remaining three quarters. Delimiting exactly where verification substitutes
for capability---and what bounds it---is the paper's result.

\paragraph{Contributions.}
\begin{enumerate}\itemsep1pt \parskip0pt
  \item \textbf{A controlled generator-substitution study at full test-set scale} under a fixed
        verification stack, establishing parity on factual entailment, a $3.3$-point completeness
        deficit against the frontier model on an axis where both far exceed the reference corpus, and
        a single deficit in structural fidelity---the axis falling furthest below the
        reference standard as well (\S\ref{sec:results}).
  \item \textbf{Section-scoped refinement}: a construction-not-instruction mechanism confining each
        revision to the sections carrying outstanding feedback, shared verbatim by both backends so
        the comparison's symmetry is a property of the code rather than a claim in the text
        (\S\ref{sec:scoped}).
  \item \textbf{Evidence that the frontier baseline is not uniformly stronger}: its provider's
        content filter refused two trauma-related transcripts outright, which on-device generation
        processed and certified---a categorical availability limit, and one of three
        constraints---with data locality and model-deprecation continuity---that determine where
        substitution is warranted (\S\ref{sec:whyslm}).
  \item \textbf{Verification, not scale, is what the small model needs.} Ablating the refinement loop
        leaves the SLM accepted on $26.8\%$ of encounters against the frontier model's $74.8\%$;
        restoring it lifts both to $83.2\%$ and $98.4\%$. The round budget is also a tunable system
        parameter, trading verification calls for autonomy and safety rather than arrangement
        (\S\ref{sec:budget}).
\end{enumerate}

\section{Related Work}
\label{sec:related}

Parameter-efficient adaptation~\cite{hu2021lora,dettmers2023qlora} has made specialization of small
open-weight models practical, and a growing literature---including a position paper for agentic
settings~\cite{belcak2025slmagentic}---argues specialized small models can displace frontier
generalists on narrow tasks. Such comparisons typically evaluate the model in isolation; we ask what
happens inside a system built to detect and correct its errors, which is how these models are
deployed. On refinement, Self-Refine~\cite{madaan2023selfrefine} improves outputs without weight
updates, but intrinsic self-correction is unreliable absent external
feedback~\cite{huang2024selfcorrect}, so we drive refinement from independent rubric-based evaluators
rather than generator introspection; verification as a hallucination-reduction mechanism is likewise
established~\cite{dhuliawala2023chainofverification}, though usually self-administered. Routing between
a cheap and an expensive model is the premise of LLM cascades and learned
routers~\cite{chen2023frugalgpt,ong2025routellm}, which escalate on a confidence score or a trained
router. Our escalation signal is different in kind: it is the same rubric-based verifier that must
certify the note for clinical use anyway, so routing is a by-product of an admissibility gate rather
than a separately calibrated component, and a kept note carries an explicit guarantee rather than a
predicted probability of adequacy. Model-swappable pipelines also introduce a hazard absent from
single-model systems: evaluators that locate structure by surface pattern encode the conventions of
whichever generator they were built against, so a substitution can shift scores for reasons unrelated
to quality---and being LLM judges, they carry known pathologies such as verbosity
bias~\cite{zheng2023llmjudge}. Prior clinical dialogue-to-note work~\cite{yim2023acibench} largely
uses frontier generators and reports reference-based quality; our setting provides both signals and
lets us observe them diverge.
% \section{Method}
% \label{sec:method}
% \begin{figure}[t]
% \centering
% \includegraphics[width=\linewidth]{design.png}
% \caption{The pipeline. Only the generator differs between conditions; evaluators, rubrics, thresholds
% and the refinement policy are shared verbatim.}
% \label{fig:pipeline}
% \end{figure}
\begin{figure}[t]
\centering
\includegraphics[width=0.8\linewidth]{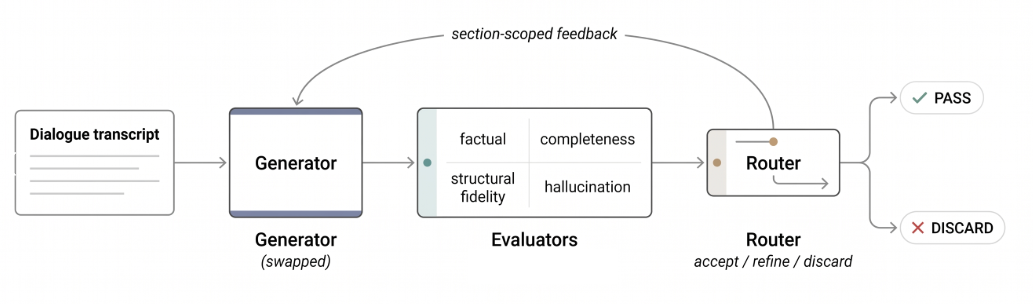}
\caption{The pipeline: only the generator differs between conditions; evaluators, rubrics, and the
refinement policy are shared verbatim.}
\label{fig:pipeline}
\end{figure}
\paragraph{Architecture.}
The \textbf{generator} produces a candidate SOAP note (Subjective, Objective, Assessment, Plan) from
the transcript---the only component swapped between conditions (Figure~\ref{fig:pipeline}).
\textbf{Evaluators} run concurrently: a \emph{factual verifier} extracts atomic claims per section
and tests each for entailment against the transcript; a \emph{completeness evaluator} tests semantic
coverage of key clinical facts across six categories (symptoms, findings, diagnoses, medications,
procedures, vitals); a \emph{structural fidelity} evaluator (SFS) scores three inter-section
consistency chains (Objective$\to$Assessment, Assessment$\to$Plan, boundary discipline) on ordinal
rubrics normalized to $[0,1]$; a \emph{hallucination categorizer} assigns a category and an
independent severity to each unsupported claim. An \textbf{aggregator/router} merges feedback into a
severity-ordered, section-annotated list, then accepts, discards, or returns the note for refinement.
All evaluators are instantiated with the same frontier judge (Claude Sonnet 4.6) in both conditions,
isolating generator quality from grading criteria; because neither generator under test is the judge,
the comparison is free of the self-preference bias documented for LLM
judges~\cite{zheng2023llmjudge}. Rubrics and formulas are in Appendix~\ref{app:rubrics}.

\paragraph{Acceptance policy.}
All four evaluators are rubric-based: fixed schemas returning structured per-item labels aggregated
deterministically, not holistic scores. Crucially, not every unsupported claim is an error. Two of the
categorizer's four categories denote ordinary documentation convention---pertinent negatives, routine
differential phrasing---rather than fabrication, so a score that pools all unsupported claims
overstates clinical risk: among validation samples falling just below threshold, a majority carried
\emph{zero} claims in the two fabrication-relevant categories. We therefore gate acceptance on
$\mathrm{factual}_{\mathrm{clin}}(n)$, which counts only fabrication-relevant claims against the note
(Appendix~\ref{app:rubrics}); it is computed in the shared decision node, so both conditions receive
it identically. A note is accepted iff
\[
\mathrm{accept}(n) =
\big[\mathrm{factual}_{\mathrm{clin}}(n)\!\geq\!0.90\big] \wedge
\big[\mathrm{completeness}(n)\!\geq\!90\%\big] \wedge
\big[\mathrm{SFS}(n)\!\geq\!0.60\big] \wedge
\big[\mathrm{crit}(n)\!=\!0\big],
\]
evaluated each round up to $N{=}3$; failing any axis at round $N$ is a hard discard.

\paragraph{Section-scoped refinement.}
\label{sec:scoped}
A refinement round conditioned on the full note invites the generator to rewrite content no evaluator
flagged---largely benign for a frontier generator, but not for the 4B model, where satisfying one
section's completeness deficiency introduced structural violations in another, including verbatim
duplication of Plan content into Assessment. Instruction-level mitigation (``preserve sections
identified as passing'') proved insufficient, consistent with the unreliability of instruction-only
self-correction~\cite{huang2024selfcorrect}, so we make preservation structural. Letting $T(F)\subseteq\mathcal{S}=\{S,O,A,P\}$ be the section target set induced by round
$k$'s aggregated feedback $F$, and $r$ the generator's revision,
\[
n_{k+1}[s] =
\begin{cases}
r[s], & s \in T(F)\ \wedge\ r[s] \neq \emptyset\\[2pt]
n_k[s], & \text{otherwise,}
\end{cases}
\qquad s \in \mathcal{S},
\]
so sections outside $T(F)$ are byte-identical across rounds by construction, and a parse failure
degrades to ``no change'' rather than an empty section. Feedback-to-section attribution becomes
load-bearing under scoping: the completeness evaluator maps every missing \emph{procedure} to Plan,
but a procedure mentioned in dialogue is often already performed and belongs in history, so we widen
that category's target set to $\{S,O,P\}$. The policy is implemented once and shared verbatim by both
backends, each supplying only a renderer for its native section-header convention; the two conventions
and an example note from each are shown in Appendix~\ref{app:examples}.

\section{Experimental Setup}
\label{sec:setup}

\textbf{Data.} A synthetic dialogue-to-SOAP corpus~\cite{omihealth2024soap,wang2024notechat} of
$9{,}250$ train / $500$ validation / $250$ test transcript--note pairs, each with a paired reference
SOAP note. All reported results are on the test split, with thresholds fixed on validation
beforehand.
\textbf{Generators.} \emph{Baseline:} GPT-4.1 via commercial API, the production frontier model for
this task at the time of the study; reported costs are specific to its published rates, and
Appendix~\ref{app:cost} gives the sensitivity to that choice. \emph{SLM:} Gemma-3-4B-IT
\cite{gemmateam2025gemma3} with a QLoRA adapter (rank $16$, $\alpha{=}32$, attention and MLP
projections) fine-tuned on the training split, served locally in bf16; full training configuration in
Appendix~\ref{app:finetune}.
\textbf{Deployment.} \emph{Generation} in the SLM condition runs entirely on one consumer machine
(Apple Silicon, 64\,GB unified memory, no discrete GPU). Verification calls a hosted judge in both
conditions, so transcripts do leave the machine for scoring: what the SLM removes is the per-encounter
\emph{generation} dependency. Keeping protected health information wholly within a local trust
boundary would additionally require serving the verifiers on-device, which is also where the
pipeline's cost sits (Appendix~\ref{app:cost}).
\textbf{Policy.} Thresholds $(0.90,\,90\%,\,0.60)$ and zero critical hallucinations; $N{=}3$;
generator temperature $0.25$ in both conditions.

\section{Results}
\label{sec:results}

\newcommand{\tbd}{\textcolor{red}{\textbf{TBD}}}
\begin{table}[t]
\centering
\small
\setlength{\tabcolsep}{4.5pt}
\caption{Generator substitution under a fixed verification stack, full test split. Both SLM columns
come from the same generations, so $N{=}3$ vs.\ $N{=}5$ isolates the round budget. All columns cover
the same $250$ encounters; GPT-4.1 produced notes for only $248$---its provider refused two
transcripts (below)---so those two count as not accepted, while its quality means are over the $248$
that exist. Latency is per round; per-note latency and generator token cost are resolved in
Appendix~\ref{app:rounds}.}
\label{tab:results}
\begin{tabular}{lrrrr}
\toprule
 & \textbf{GPT-4.1} & \multicolumn{2}{c}{\textbf{Gemma-3-4B + QLoRA}} & \textbf{Reference} \\
\cmidrule(lr){3-4}
\textbf{Metric} & ($N{=}3$) & ($N{=}3$) & ($N{=}5$) & \textbf{notes} \\
 & $n{=}250$ & $n{=}250$ & $n{=}250$ & $n{=}250$ \\
\midrule
Factual accuracy, clinical (mean) & \textbf{0.992} & 0.984 & 0.986 & 0.989 \\
Completeness (mean)               & 98.5\% & 95.2\% & 95.7\% & 82.5\% \\
Structural fidelity, SFS (mean)   & 0.843 & 0.778 & 0.777 & \textbf{0.873} \\
Critical hallucinations (total)   & 1 & 13 & \textbf{6} & 10 \\
Verifier-accepted rate            & \textbf{98.4\%} & 75.2\% & 83.2\% & 32.4\% \\
\midrule
Mean refinement rounds            & 1.28 & 2.07 & 2.53 & --- \\
Median latency per round          & \textbf{57.5}\,s & 81.3\,s & 85.0\,s & --- \\
Generation locality               & third-party API & \textbf{on-device} & \textbf{on-device} & --- \\
\bottomrule
\end{tabular}
\end{table}

\paragraph{How strict is the acceptance gate? A reference-note calibration check.}
\label{sec:calibration}
An acceptance rate is interpretable only relative to the standard it encodes, so we scored the
corpus's own reference notes through the identical evaluator stack and predicate
(Table~\ref{tab:results}, right column). On the axes governing clinical admissibility the gate tracks
the corpus: reference notes clear the factual threshold in $99\%$ of cases and structural fidelity in
$96\%$, and their SFS mean ($0.873$) exceeds both generators'---gold $>$ frontier $>$ SLM, the
ordering a valid metric should produce. The completeness threshold sits above the corpus's own level:
reference notes average $82.5\%$ against a $90\%$ bar and fail it in $66\%$ of cases, so the composite
gate admits only $32.4\%$ of the reference corpus. On that axis both generators therefore exceed the
corpus rather than falling short of an absolute standard, and acceptance should be read as conformance
to a deliberately conservative gate. The remaining gap is not a threshold artifact: relaxing completeness
to the reference mean raises SLM autonomy only to $84.8\%$ and an implausibly permissive $70\%$ bar
reaches $90.4\%$, so what persists is structural-fidelity and critical-hallucination failure. The verifier
also flags critical claims in $9$ of the reference notes, holding the corpus to the standard it applies
to our generators (Appendix~\ref{app:calib}).
\paragraph{The deficit is isolated to arrangement, not content.}
Claim-level factual accuracy is $0.984$ against $0.992$---a $0.008$ gap, detectable on paired scores
($n{=}248$, Wilcoxon $p{=}1\times10^{-4}$) but amounting to one additional unsupported claim per
$125$ on the axis governing clinical admissibility.
Completeness is $95.2\%$ against $98.5\%$, and although that $3.3$-point gap is implicated in $47\%$
of the SLM's rejections, both generators sit far above the reference corpus ($82.5\%$) and the
frontier model's excess over it ($+16.0$ points) is the larger of the two---partly earned by
explicitly marking information the transcript does not contain (Appendix~\ref{app:examples}).
Structural fidelity is the exception: the axis on which the SLM falls furthest below \emph{both} the
frontier model and the reference notes ($0.778$ against $0.843$ and $0.873$), and the one implicated
in $56\%$ of its $62$ rejections against factual accuracy's $5\%$. The calibration above separates the
two leading categories: the corpus clears the SFS threshold but fails completeness, so a completeness
rejection indicts a bar the corpus itself does not meet while an SFS rejection indicts one it clears.
That asymmetry is why we localize the deficit to arrangement, and it is the more tractable
failure---the model has the right facts in the wrong place, which section-scoped refinement attacks
structurally (\S\ref{sec:scoped}) and a larger round budget partially resolves
(Table~\ref{tab:ablation}). Critical hallucinations are higher in count
($13$ vs.\ $1$), but zero of them is one of the four acceptance conditions, so all $13$ fall in the
escalated set and none appears in accepted output by construction.

\paragraph{The verifier turns the gap into a routing decision, and the routed system loses nothing.}
\label{sec:routing}
We call the fraction of encounters the SLM completes without escalation its \emph{autonomy rate}---its
verifier-accepted rate in Table~\ref{tab:results}---at $75.2\%$, against $98.4\%$ for the frontier
generator. What matters is that the shortfall is \emph{reported}: every rejected note is caught by
verification rather than surfaced, so what the SLM keeps passes the same verifier-defined
admissibility criterion as the frontier condition's, and the resulting policy can be measured end to end. It accepts $249/250$ ($99.6\%$)
against frontier-only's $246/250$ ($98.4\%$)---three discordant encounters, all favouring routing,
statistically indistinguishable (exact McNemar $p{=}0.25$)---and well above the SLM alone
($188/250$), while sending only the $62$ rejected encounters to the frontier model ($90$ requests
against $318$). Frontier \emph{generation} spend falls $3.4\times$, \$$2.50$ to \$$0.73$, though
verification dominates total spend in both conditions (Appendix~\ref{app:cost}). The one residual
failure is a transcript both generators reject, so no policy over these two models could have admitted
it; conversely, both transcripts the content filter refused are notes the SLM keeps, which the routed
system never submits to the filter. Without a per-note admissibility signal, a small model's failures
would be indistinguishable from its successes.
\paragraph{Where substitution is warranted: the baseline is not uniformly stronger.}
\label{sec:whyslm}
GPT-4.1 produced no note for two encounters---both rejected with a content-management-policy error on
the \emph{prompt}, concerning trauma and self-harm---so they count against its acceptance rate rather
than leaving the denominator. The on-device SLM, subject to no such gate, processed both and its
verifier accepted both on the first round at factual $1.000$ with no critical hallucinations. At
$2/250$ this is not an estimate of the deployment rate, since refusals are content-dependent, but it is
the first of three properties motivating substitution that no quality metric expresses.
\emph{Availability}: a declined transcript cannot be documented at any price or latency.
\emph{Locality}: three quarters of encounters complete with no third-party generation call, a
constraint rather than a preference in regulated settings. \emph{Continuity}: a hosted generator's
price, availability and behaviour are the provider's to change, whereas a served adapter is a fixed
artifact. All three concern the generation path; verification retains a hosted dependency
(\S\ref{sec:setup}), though the risk is generation-specific rather than inherent to hosted
inference---the judge scored both refused transcripts on all four axes without degradation---and
because a judge must discriminate rather than generate, relocating it is the tractable next step
(Appendix~\ref{app:cost}). Where these constraints bind, the question is not which generator is more accurate but whether a given note meets the system’s verifier-defined admissibility criteria; the verifier evaluates that note by note.
\begin{table}[t]
\centering
\small
\setlength{\tabcolsep}{5pt}
\caption{Ablating the refinement loop for both generators over the same $250$ encounters. Budgets are
derived by truncating one run per generator, so within a generator only the cutoff varies; $N{=}1$ is a
single generation pass scored once by the gate. The frontier run had a ceiling of $N{=}3$, the SLM run
$N{=}5$. Reference agreement is against the corpus note, which no evaluator consumes; the SLM's higher
values there follow from fine-tuning on this corpus and are not evidence of better writing.}
\label{tab:ablation}
\begin{tabular}{lrrrr}
\toprule
& \textbf{Accepted} & \textbf{ROUGE-1} & \textbf{ROUGE-L} & \textbf{BLEU} \\
\midrule
\multicolumn{5}{l}{\emph{Gemma-3-4B + QLoRA} (on-device)} \\
\quad $N{=}1$ (no refinement) & \textbf{26.8\%} & 0.709 & 0.536 & 41.8 \\
\quad $N{=}2$ & 66.0\% & 0.695 & 0.511 & 36.7 \\
\quad $N{=}3$ & 75.2\% & 0.686 & 0.503 & 33.4 \\
\quad $N{=}4$ & 79.2\% & 0.688 & 0.504 & 34.5 \\
\quad $N{=}5$ & \textbf{83.2\%} & 0.686 & 0.502 & 32.3 \\
\midrule
\multicolumn{5}{l}{\emph{GPT-4.1} (frontier API)} \\
\quad $N{=}1$ (no refinement) & \textbf{74.8\%} & 0.568 & 0.381 & 18.2 \\
\quad $N{=}2$ & 95.6\% & 0.566 & 0.379 & 18.0 \\
\quad $N{=}3$ & \textbf{98.4\%} & 0.566 & 0.379 & 17.9 \\
\bottomrule
\end{tabular}
\end{table}

\paragraph{Ablation: verification is what the small model needs.}
\label{sec:budget}
Table~\ref{tab:ablation} ablates the loop by truncating the round budget. Stripped of refinement the
SLM clears the gate on $26.8\%$ of encounters against the frontier model's $74.8\%$; restoring it
lifts the SLM to $83.2\%$ and the frontier model to $98.4\%$. The loop therefore closes most of a
$48$-point deficit for the small model while adding $23.6$ points for the large one, which needs
$1.28$ rounds on average---verification does the most work exactly where the generator is weakest,
which is the substitution this paper is about. The gains are axis-specific: over that range critical
hallucinations fall from $26$ to $6$ and completeness climbs $12.7$ points, while structural fidelity
does not improve ($0.803\rightarrow0.777$), so refinement buys autonomy and safety rather than
arrangement (Figure~\ref{fig:budget}). Because rounds \emph{generate} on-device the budget is a dial
rather than a generation-cost constraint: $N{=}3$ to $N{=}5$ adds no generation spend and $22\%$ more
verification calls (Appendix~\ref{app:cost}). The truncation also validates the comparison---the
$N{=}3$ figure derived this way ($75.2\%$) matches an independent $N{=}3$ run over the same encounters
($74.4\%$) to within $0.8$ points.

\paragraph{Acceptance cannot validate the loop, so we test the gate against a metric outside it.}
Acceptance is the objective the loop optimizes, so Table~\ref{tab:ablation} also reports agreement with
the corpus reference note, which no evaluator consumes. Two things follow. First, refinement moves
\emph{away} from it---ROUGE-L falls $0.536\rightarrow0.502$, attributable to revision rather than
selection, because the completeness rubric rewards enumeration the reference notes do not supply
(Appendix~\ref{app:drift})---so refinement buys admissibility under a conservative rubric, not fidelity
to the corpus's style. Second, and this is the property routing relies on, the gate nonetheless
\emph{discriminates} on that same external signal: kept notes agree with the reference substantially
more closely than escalated ones, ROUGE-1 $0.700$ vs.\ $0.646$ and ROUGE-L $0.518$ vs.\ $0.456$, both
significant under a one-sided Mann--Whitney $U$ test ($p{=}4.5\times10^{-3}$ and
$p{=}1.6\times10^{-4}$). Its decisions therefore track a dimension it was never tuned against even
where its revisions drift from it. Within the axes it does score the separation is starker: structural
fidelity $0.828$ vs.\ $0.627$, and all $13$ critical hallucinations fall in the escalated set. Kept
notes score $0.987$ on factual accuracy, against $0.989$ for the reference corpus itself
(Appendix~\ref{app:separation}).
\section{Limitations and Future Work}
\label{sec:limitations}

The acceptance gate certifies conformance to our rubric-based verifier, not independently validated
clinical correctness; the reference-note calibration (\S\ref{sec:calibration}) and the
kept-versus-escalated separation (Appendix~\ref{app:separation}) are indirect checks on that
verifier, not a clinician-adjudicated ground truth or a formal safety guarantee. Our results are on
a synthetic dialogue corpus~\cite{omihealth2024soap}, so replication on real
encounters with clinician-authored notes is the natural next step; the reference comparisons here
establish divergence from an independent reference standard rather than from clinical documentation
practice specifically. Whether other small models reach a comparable share of the
workload, and how the approach behaves on less structured generation tasks, is likewise open; the
verification stack itself requires only a section-header renderer per backend
(\S\ref{sec:scoped}). Serving the evaluators with a small model is the direction we would pursue next:
it addresses the residual external dependency and the dominant cost line at once
(Appendix~\ref{app:cost}), and how far judge capability can be reduced before the gate loses the
separation of Appendix~\ref{app:separation} is an empirical question we have not answered.

\section{Conclusion}
Inside a fixed verification architecture, a 4B-parameter generator shows a $0.008$ absolute gap in claim-level factual accuracy relative to a frontier model and a 3.3-point gap in completeness, with the residual deficit concentrated in structural fidelity: the right clinical facts in the wrong section.
% Inside a fixed verification architecture, a 4B-parameter generator holds claim-level factual accuracy
% to within $0.008$ of a frontier model and completeness to within $3.3$ points, with the residual
% deficit confined to structural fidelity: the right clinical facts in the wrong section. 
Ablating the loop separates architecture from scale---unrefined, the small model clears the gate on $26.8\%$ of
encounters against $83.2\%$ with refinement---showing that, in this evaluated setting, verification and refinement substantially improve the small model's verifier-accepted rate.
Because the shortfall is an explicit verifier decision it becomes a routing policy that yields verifier-accepted outputs for 99.6\% of encounters, compared with 98.4\% for the frontier-only condition while removing the frontier
\emph{generation} dependency for three quarters
of encounters, and the comparison is not one-directional: in this evaluation, the hosted frontier service returned no note for two trauma-related transcripts because of provider content-management-policy errors, while the locally served small model generated verifier-accepted outputs for both. These two cases illustrate a provider-controlled availability constraint observed in the hosted generation path, rather than a general limitation of the frontier model.
For structured, source-grounded generation, a per-note
admissibility gate is what makes substitution measurable. 
It turns a capability gap into a routing decision, and the fraction of source-grounded clinical-document generation tasks handled by the small model becomes a measured quantity rather than an assumption.

\bibliographystyle{plain}
\bibliography{references}

\appendix

\section{Round-budget ablation}
\label{app:budget}

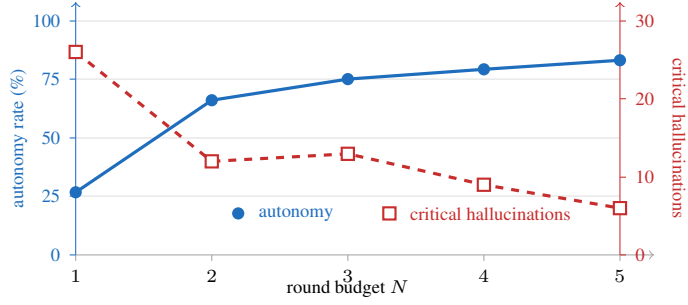
\begin{figure}[h]
\centering
\begin{tikzpicture}[x=1cm,y=1cm]
  \definecolor{autoblue}{RGB}{31,110,190}
  \definecolor{critred}{RGB}{200,45,45}
  % gridlines
  \draw[gray!25] (0,0.78)--(7.2,0.78) (0,1.55)--(7.2,1.55) (0,2.33)--(7.2,2.33) (0,3.10)--(7.2,3.10);
  \draw[->,gray!70] (0,0)--(7.65,0);
  \draw[->,autoblue] (0,0)--(0,3.35);
  \draw[->,critred] (7.2,0)--(7.2,3.35);
  \foreach \x/\n in {0/1, 1.8/2, 3.6/3, 5.4/4, 7.2/5}
    \draw[gray!70] (\x,0)--(\x,-0.08) node[below,font=\scriptsize,black] {$\n$};
  \foreach \y/\v in {0/0, 0.78/25, 1.55/50, 2.33/75, 3.10/100}
    \draw[autoblue] (0,\y)--(-0.08,\y) node[left,font=\scriptsize] {\v};
  \foreach \y/\v in {0/0, 1.03/10, 2.07/20, 3.10/30}
    \draw[critred] (7.2,\y)--(7.28,\y) node[right,font=\scriptsize] {\v};
  % autonomy: rises (blue, solid, circles)
  \draw[very thick,autoblue] (0,0.83)--(1.8,2.05)--(3.6,2.33)--(5.4,2.46)--(7.2,2.58);
  \foreach \p in {(0,0.83),(1.8,2.05),(3.6,2.33),(5.4,2.46),(7.2,2.58)}
    \fill[autoblue] \p circle (2.3pt);
  % critical hallucinations: fall (red, dashed, squares)
  \draw[very thick,critred,dashed] (0,2.69)--(1.8,1.24)--(3.6,1.34)--(5.4,0.93)--(7.2,0.62);
  \foreach \p in {(0,2.69),(1.8,1.24),(3.6,1.34),(5.4,0.93),(7.2,0.62)}
    \draw[critred,fill=white,thick] \p +(-2.3pt,-2.3pt) rectangle +(2.3pt,2.3pt);
  % axis labels, coloured to match their series
  \node[rotate=90,font=\scriptsize,autoblue,align=center] at (-0.75,1.55) {autonomy rate (\%)};
  \node[rotate=-90,font=\scriptsize,critred,align=center] at (7.98,1.55) {critical hallucinations};
  \node[font=\scriptsize] at (3.6,-0.45) {round budget $N$};
  % inline legend
  \fill[autoblue] (2.15,0.55) circle (2.3pt);
  \node[font=\scriptsize,anchor=west,autoblue] at (2.3,0.55) {autonomy};
  \draw[critred,fill=white,thick] (4.15,0.55) +(-2.3pt,-2.3pt) rectangle +(2.3pt,2.3pt);
  \node[font=\scriptsize,anchor=west,critred] at (4.3,0.55) {critical hallucinations};
\end{tikzpicture}
\caption{Raising the round budget buys autonomy \emph{and} safety on identical generations: autonomy
rises $26.8\rightarrow83.2\%$ while critical hallucinations fall $26\rightarrow6$ ($N{=}2\rightarrow3$
is the one non-monotonic step), with gains not yet plateaued at $N{=}5$. Every budget is derived from
one $N{=}5$ run by truncation, so only the cutoff varies.}
\label{fig:budget}
\end{figure}

Figure~\ref{fig:budget} plots the two safety-relevant series of Table~\ref{tab:ablation} for the SLM.

\section{Evaluator rubrics and scoring formulas}
\label{app:rubrics}

Claims are extracted \emph{per section}: for each $s\in\mathcal{S}=\{S,O,A,P\}$ an extraction prompt
returns a claim set $C_s$ (with a stricter prompt for $P$ excluding forward-looking rationale and
boilerplate); each $c\in C_s$ is tested against transcript $\tau$ by a separate judge call returning
binary $\mathrm{ent}(c)$. The reported score is a pooled micro-average, not a mean of per-section
scores---so a note with one claim-dense section and three sparse ones is scored by the dense section:
\[
\mathrm{factual}(n) = \frac{\sum_{s}\sum_{c\in C_s}\mathrm{ent}(c)}{\sum_{s}|C_s|}.
\]
\[
\mathrm{factual}_{\mathrm{clin}}(n) = \frac{\sum_{s}|C_s| - \big|\{c : \mathrm{ent}(c)=0,\
\mathrm{cat}(c)\in\{\textsc{Inaccurate\_Data},\textsc{Needs\_Review}\}\}\big|}{\sum_{s}|C_s|}.
\]
Completeness pools counts across the six categories rather than averaging category percentages:
$\mathrm{completeness}(n) = 100\times\sum_k\sum_{f\in K_k}\mathrm{pres}(f)\,/\,\sum_k|K_k|$, with
$\mathrm{pres}(f)$ testing semantic, not exact-string, presence. Structural fidelity scores each chain
$\mathrm{ch}\in\{\mathrm{OA},\mathrm{AP},\mathrm{B}\}$ on an ordinal rubric
$r_{\mathrm{ch}}\in\{0,1,2,3\}$ with fixed qualitative anchors at every level (the boundary chain
defines $3$ as ``clean boundaries throughout, no cross-section bleed'' down to $0$ as ``section
collapse''), renormalized over the chains $V$ whose judge call succeeded:
$\mathrm{SFS}(n) = \frac{1}{3|V|}\sum_{\mathrm{ch}\in V} r_{\mathrm{ch}}(n)$. The hallucination
categorizer operates only on claims with $\mathrm{ent}(c)=0$, assigning a category from a fixed
four-way taxonomy and, \emph{independently}, a severity---severity is not derived from category by any
hard-coded rule---with $\mathrm{crit}(n) = |\{c : \mathrm{ent}(c)=0,\ \sigma(c)=\textsc{Critical}\}|$
over all categories.

\paragraph{Threshold defensibility.} SFS is discretized to ninths, so no sample in a $100$-sample
validation sweep scores between $5/9$ and $6/9$: $\theta_s{=}0.60$ is already equivalent to requiring
$\geq 6/9$. Inspecting the $10$ samples at exactly $5/9$ confirmed substantive violations (diagnostic
interpretation contaminating Objective; generic Plan content despite Assessment naming specific
diagnoses) rather than borderline near-misses, so we did not lower it.

\section{Generator cost, resolved by round type}
\label{app:rounds}

Table~\ref{tab:results} reports generator cost per \emph{call}. This appendix gives the full
decomposition, including the per-note totals those per-call figures compose into, so that no quantity
in the main text has to be inferred.

Two facts govern the whole table. First, a refinement prompt costs $\approx\!1.9\times$ a fresh
prompt---it re-sends the transcript alongside the previous note, the outstanding feedback items and the
positive anchors---and it does so almost identically in both conditions ($1.88\times$ for the frontier
generator, $1.89\times$ for the SLM). Second, the prompt text itself is shared: both generators receive
byte-identical system prompts and user-message templates, so per-call cost differs only by tokenizer
($1.02$--$1.03\times$, Gemma's vocabulary being marginally less efficient on this text). Consequently
the per-note difference between the two systems is \emph{entirely} a difference in how many calls each
makes, not in what a call costs: the SLM's rounds are $53\%$ refinement against the frontier model's
$22\%$, which is also why a single blended per-round average would misrepresent per-call cost as
$1.26\times$ rather than $1.02\times$.

\begin{table}[h]
\centering
\small
\setlength{\tabcolsep}{4pt}
\caption{Generator token cost by round type, means over the test split. ``Fresh'' is the round-$0$
generation from the transcript; ``refinement'' is every subsequent round. Per-note columns sum over all
of a note's rounds. Each row is counted with its own model's tokenizer, so counts are comparable down a
column but not across the two generators as a single unit.}
\label{tab:roundtype}
\begin{tabular}{lrrrrrrrrr}
\toprule
& & \multicolumn{3}{c}{\textbf{Fresh call}} & \multicolumn{3}{c}{\textbf{Refinement call}}
& \multicolumn{2}{c}{\textbf{Per note}} \\
\cmidrule(lr){3-5}\cmidrule(lr){6-8}\cmidrule(lr){9-10}
\textbf{Condition} & \textbf{rnds} & \textbf{calls} & \textbf{in} & \textbf{out}
& \textbf{calls} & \textbf{in} & \textbf{out} & \textbf{in} & \textbf{out} \\
\midrule
GPT-4.1 $N{=}3$ & 1.28 & 248 & 1{,}415 & 547 & 70 & 2{,}658 & 603 & 2{,}166 & 717 \\
Gemma $N{=}3$   & 2.07 & 250 & 1{,}450 & 344 & 281 & 2{,}744 & 466 & 4{,}534 & 867 \\
Gemma $N{=}5$   & 2.53 & 250 & 1{,}450 & 347 & 382 & 2{,}810 & 517 & 5{,}743 & 1{,}138 \\
\bottomrule
\end{tabular}
\end{table}

The same decomposition applies to latency. Per note the SLM takes a median $171.6$\,s against
$60.5$\,s, a $2.8\times$ gap; per round it is $81.3$\,s against $57.5$\,s. Restricting to the
$n{=}76$ encounters on which both generators happened to use the same number of rounds---which
controls for difficulty as well as round count---gives $81.2$\,s against $59.9$\,s, a $1.36\times$
per-round gap. The residual per-round difference is generation throughput; the per-note difference is
overwhelmingly the round count, which is a policy parameter rather than a property of the model
(\S\ref{sec:budget}).

\section{Kept-versus-escalated separation}
\label{app:separation}

Full per-axis separation behind the validation result in \S\ref{sec:results}.

\begin{table}[h]
\centering
\small
\caption{What the verifier separates. Rows 1--4 are the axes the gate itself scores; rows 5--6 are
lexical agreement with the corpus reference note, a signal \emph{no} in-loop evaluator optimizes.
ROUGE is undefined for the reference column, which is the comparison target. Fine-tuning on this
corpus's reference distribution inflates the SLM's absolute ROUGE, but affects both kept and escalated
notes identically, so the kept--escalated contrast---not the absolute value---is the claim.}
\label{tab:separation}
\begin{tabular}{lrrr}
\toprule
\textbf{Metric} & \textbf{Kept} ($n{=}188$) & \textbf{Escalated} ($n{=}62$) & \textbf{Reference} ($n{=}250$) \\
\midrule
Factual accuracy, clinical      & 0.987 & 0.975 & 0.989 \\
Completeness                    & 96.5\% & 91.2\% & 82.5\% \\
Structural fidelity, SFS        & \textbf{0.828} & \textbf{0.627} & 0.873 \\
Critical hallucinations (total) & \textbf{0} & \textbf{13} & 10 \\
\midrule
ROUGE-1 F1 vs.\ reference       & \textbf{0.700} & \textbf{0.646} & --- \\
ROUGE-L F1 vs.\ reference       & \textbf{0.518} & \textbf{0.456} & --- \\
\bottomrule
\end{tabular}
\end{table}

\section{Reference-note calibration: additional detail}
\label{app:calib}

The critical-hallucination flags on reference notes are worth stating precisely, since a detector
tuned to favour our own generator would be a fatal circularity. Our verifier flags unsupported
critical claims in $10$ instances across $9$ of the $250$ reference notes, the majority in the
fabrication-relevant category---including a medication appearing nowhere in the source dialogue.
Severity is assigned independently of category, so a minority of these flags fall in categories the
entailment criterion treats as documentation convention rather than fabrication.

Part of the frontier model's acceptance advantage is closer conformance to an exhaustiveness rubric
the reference notes themselves do not satisfy: GPT-4.1's completeness ($98.5\%$) is \emph{further}
from the reference mean ($82.5\%$) than the SLM's ($95.2\%$). This is the verbosity bias documented
for LLM judges~\cite{zheng2023llmjudge} surfacing in a clinical rubric, and it is the reason we treat
the completeness axis as measuring conformance to our rubric rather than clinical adequacy.

\section{Refinement drifts from the reference while improving the judge's metric}
\label{app:drift}

Across all $n{=}183$ multi-round samples, LLM-judged completeness improved in $176$ (mean $+17.3$
points) while ROUGE-L against the paired reference note---a signal no evaluator optimizes---worsened
in $165$. The aggregate effect is visible down the reference-agreement columns of
Table~\ref{tab:ablation}, and it decomposes into selection and revision: notes that were never refined
score $0.718$ ROUGE-1 against $0.705$ for notes that would go on to be refined, measured \emph{before}
any revision, so selection accounts for only $0.013$ of the decline while revision itself accounts for
$-0.031$. The mechanism is a definitional mismatch: the completeness rubric rewards exhaustive
enumeration of transcript facts, whereas the corpus's reference notes are markedly more selective, so
the loop reports monotone improvement while drifting from the artifact it is nominally imitating. The
SLM's higher lexical overlap should not be read as better writing---it was fine-tuned on this corpus's
own reference distribution---but the divergence \emph{direction} is the point: where references exist,
a metric that no component optimizes is worth tracking as a guardrail rather than reported post hoc.

\section{End-to-end cost accounting}
\label{app:cost}

An agentic pipeline's economics are set by its verification traffic rather than its generator, so we
account for both lines over the $250$-encounter test split.

\paragraph{Call structure.} Each refinement round issues one generation request and a fixed set of
judge requests: four claim-extraction calls (one per SOAP section), four claim-verification calls, one
completeness call, and three structural-fidelity chain calls---twelve per round---plus one
categorization call per unsupported claim. Measured over the split this is $13.4$ judge calls per round
in the frontier condition and $14.9$ in the SLM condition. Because the claim-verification,
completeness and categorization prompts each carry the full dialogue transcript, judge \emph{input}
volume is an order of magnitude above generation input: a median $9{,}779$ tokens per round against
$1{,}450$ for a fresh generation.

\begin{table}[h]
\centering
\small
\caption{Cost over the $250$-encounter test split at public list rates: GPT-4.1 at \$2/\$8 per million
input/output tokens, hosted judge at \$3/\$15. Generation is billed only in the frontier condition,
since SLM generation is on-device; verification is billed in both. Requests exceed routed encounters
because a note may be refined over several rounds: the $62$ escalated encounters take the frontier
model $1.45$ rounds each on average, hence $90$ generation requests.}
\label{tab:cost}
\begin{tabular}{lrr}
\toprule
& \textbf{Frontier only} & \textbf{SLM + escalation} \\
\midrule
Encounters routed to the frontier model & 250 & \textbf{62} \\
Frontier generation requests & 318 & \textbf{90} \\
\quad generation cost        & \$2.50 & \textbf{\$0.73} \\
\midrule
Rounds executed              & 318 & 608 \\
Judge calls                  & \textbf{4{,}246} & 9{,}055 \\
Judge input tokens           & \textbf{3.62}\,M & 5.45\,M \\
\quad verification cost      & \textbf{\$26.78} & \$50.31 \\
\midrule
\textbf{Total}               & \textbf{\$29.28} & \$51.04 \\
\textbf{Per note}            & \textbf{\$0.117} & \$0.204 \\
\bottomrule
\end{tabular}
\end{table}

Verification accounts for $91\%$ of spend in the frontier condition and $99\%$ in the routed one, and
is billed identically per round in both. It is therefore best read as the fixed price of the
admissibility guarantee rather than as a property of either generator: a deployment that wants notes
certified before a clinician sees them pays it whichever model writes them. Substituting the generator
changes two things only---the generation line, which falls $3.4\times$, and the number of rounds, which
rises. The second dominates, so the routed system costs $1.74\times$ more in total. What routing buys
is the elimination of a per-encounter frontier \emph{generation} call for three quarters of the
workload, together with the safety and availability properties in \S\ref{sec:results}; it is not a
cost optimization.

\paragraph{Measurement basis.} Call counts and rounds are taken from the run logs. Judge input tokens
are reconstructed by rebuilding each evaluator prompt from the same prompt constants the pipeline uses,
with the logged note, transcript and unsupported-claim counts, then tokenising with a byte-level BPE
tokenizer as a proxy for the judge's own vocabulary. Judge output is taken at $250$ tokens per call,
which is representative of the short structured verdicts these rubrics return; output is $58\%$ of the
verification line, so a $\pm50\%$ error there moves the per-note totals by roughly $\pm30\%$ and does
not affect the ordering between conditions.

\paragraph{The ordering is robust to price, and the lever is the judge.} Re-pricing our measured token
volumes against a generator costing $2.5\times$ more per input token and $3.75\times$ more per output
token raises generation's share of spend from $9\%$ to $23\%$ and narrows the ratio between conditions
from $1.74\times$ to $1.51\times$---it does not reverse it. Reversal would require a generator above
roughly \$$26$ per million input tokens, an order of magnitude beyond current chat-tier pricing, because
routing carries a fixed verification penalty of \$$23.53$ from executing more rounds. Re-pricing the
\emph{judge} instead moves the total by tens of percent, so the effective cost lever is the evaluator
stack. Two consequences follow. The configuration in which routing reduces total spend is one pairing
an expensive generator with a cheaper judge, which is the direction hosted pricing has been moving as
reasoning-tier generators are listed above general-purpose judges. And a smaller or locally-served
judge reduces total spend further than any change of generator while removing the last per-encounter
external dependency (\S\ref{sec:setup}); at published rates, moving from the judge we used to a
lower-tier model of the same family reduces total spend by $30$--$61\%$. How far judge capability can be reduced before the acceptance gate loses
discriminative power is the natural continuation of this work.

\paragraph{Configurations that reverse the ordering.} The ordering between conditions follows from the
generator's price relative to the judge's, so it is not fixed by the architecture. On our measured token
volumes neither lever alone reverses it---moving the judge to \$$1$/\$$5$ per million input/output tokens
leaves the ratio at $1.53\times$, and raising the generator to \$$10$/\$$50$ while keeping our judge
leaves it at $1.33\times$---but the pairing does: a reasoning-tier generator at \$$10$/\$$50$ with a
lower-tier judge at \$$1$/\$$5$ places the routed system at $0.90\times$ frontier-only spend (\$$20.91$
against \$$23.19$), the crossover falling near \$$7.75$ per million generator input tokens. These are
arithmetic on measured volumes rather than experiments. The substantive question they raise is how far
judge capability can be reduced while the gate retains the separation of
Appendix~\ref{app:separation}, which we leave to future work.

\section{Fine-tuning configuration}
\label{app:finetune}

The SLM adapter was trained with QLoRA parameter-efficient supervised fine-tuning on the $9{,}250$-pair
training split. Examples were formatted with Gemma's chat template as a clinical-scribe instruction plus
the dialogue, with the paired reference note as the target response. Loss was computed over the response
tokens only: each prompt--response sequence was tokenized and all prompt-side labels masked, so the
objective never rewards reproducing the instruction. Sequences were truncated at $2{,}048$ tokens.

The base model was loaded in $4$-bit NF4 with double quantization and bfloat16 compute; LoRA adapters
were applied to the attention and MLP projections (\texttt{q\_proj}, \texttt{k\_proj}, \texttt{v\_proj},
\texttt{o\_proj}, \texttt{gate\_proj}, \texttt{up\_proj}, \texttt{down\_proj}) with rank $16$,
$\alpha{=}32$ and dropout $0.05$. Training ran for $3$ epochs at per-device batch size $1$ with
$32$-step gradient accumulation, gradient checkpointing, paged AdamW ($32$-bit), cosine schedule,
learning rate $2\times10^{-4}$ and warmup ratio $0.03$. Validation loss was evaluated and checkpoints
written every $100$ steps, and the checkpoint with the lowest validation loss was selected. Note that
quantization applies to training only: at inference the base model is served in bfloat16 with the
adapter applied, as reported in \S\ref{sec:setup}.

\section{Example notes}
\label{app:examples}

Both notes below are for the same encounter (\texttt{test\_0080}) and both were accepted at round $0$.
Scores: SLM factual $1.000$, completeness $95.2$, SFS $1.000$; frontier model factual $1.000$,
completeness $100.0$, SFS $1.000$.

\paragraph{Two differences are worth reading for.} First, section headers. The system prompt requests
\texttt{**Subjective**}-style headers and the frontier model produces them in all $248$ notes it
generated; the SLM instead emits \texttt{S:/O:/A:/P:} in all $250$, which is the convention used by the
corpus's own reference notes. Supervised fine-tuning anchors surface form more strongly than an
inference-time instruction does, so the adapter reproduces the format it was trained on rather than the
one the prompt asks for. This is a concrete instance of the effect that motivates section-scoped
refinement (\S\ref{sec:scoped}): instruction-level control over form is unreliable, which is why
preservation there is enforced by construction. It does not affect the comparison, because each
backend supplies a renderer for its own header convention and every evaluator consumes the parsed
sections rather than the raw string.

Second, treatment of absent information. The frontier model enumerates fields the transcript does not
cover---\emph{Past Medical History: Not documented}, and three more in this example---averaging $3.44$
such placeholders per note against the SLM's $0.00$. Part of its completeness advantage is therefore
earned by explicitly marking absences, which is the verbosity-bias effect noted
in~\S\ref{sec:related} and the reason its excess over the reference corpus ($+16.0$ points) exceeds
the SLM's ($+12.7$).

\paragraph{SLM (Gemma-3-4B + QLoRA).}
\begin{footnotesize}
\begin{verbatim}
S: The patient reports experiencing painful right upper thoracic radiculopathy for 18
  months, with a pain intensity of 7-8 out of 10. The patient finds relief with
  nonsteroidal anti-inflammatory drugs and describes the pain as radiating from about
  3 cm from the thoracic midline on the right side. No muscle contracture or pain on
  palpation is reported.
O: Physical examination revealed no muscle contracture or pain on palpation.
  Radiographs showed no osteolysis or deformity. MRI findings included T2 hypersignal
  and T1 hyposignal at the medial side of the right T1 vertebra, with inflammation of
  the right T1 root. Radioisotope bone scanning indicated lateral hyperfixation on the
  right side of the T1 vertebra. CT scan revealed an osteolytic lesion surrounded by
  sclerotic bone at the medial wall of the right T1 pedicle, close to the dura mater
  and T1 spinal nerve.
A: The primary diagnosis is osteoid osteoma at the medial wall of the right T1
  pedicle, causing radiculopathy. Differential diagnoses could include other causes of
  osteolysis or spinal compression syndromes, but imaging and clinical presentation
  support the primary diagnosis.
P: The management plan includes surgical resection of the osteoid osteoma using a
  posterior approach with 3D imaging guidance, avoiding radiofrequency ablation due to
  proximity to the dura mater. The surgery will be performed under general anesthesia
  in a prone position using a mini-open posterior approach at the C7-T2 levels with
  the aid of Surgivisio® for vertebral level control. Post-operative care and
  follow-up will be scheduled to monitor recovery and ensure no recurrence.
\end{verbatim}
\end{footnotesize}

\paragraph{Frontier model (GPT-4.1).}
\begin{footnotesize}
\begin{verbatim}
**Subjective**

- Chief Complaint: "Painful right upper thoracic radiculopathy."
- History of Present Illness: The patient reports right upper thoracic radicular pain
  for approximately 18 months, rated as 7–8 out of 10 in severity. Pain is located
  about 3 cm from the thoracic midline on the right side, with intercostal
  irradiation. NSAIDs provide relief. No muscle contracture or pain on palpation.
- Past Medical History: Not documented.
- Medications: Nonsteroidal anti-inflammatory drugs (CONTINUING).
- Allergies: Not documented.
- Social/Family History: Not documented.
- Review of Systems: Positive for right upper thoracic pain with radiculopathy;
  negative for muscle contracture and pain on palpation.

**Objective**

- Vital Signs: Not documented.
- Physical Examination: No muscle contracture; no pain on palpation.
- Diagnostic Results:
  - Radiographs: No osteolysis or deformity.
  - MRI: T2 hypersignal and T1 hyposignal at the medial side of right T1 vertebra;
    inflammation of right T1 root.
  - Radioisotope bone scan: Lateral hyperfixation on right side of T1 vertebra.
  - CT scan: Osteolytic lesion surrounded by sclerotic bone at medial wall of right T1
    pedicle, close to dura mater and T1 spinal nerve.

**Assessment**

- Osteoid osteoma of the medial wall of the right T1 pedicle, causing right upper
  thoracic radiculopathy and inflammation of the right T1 root. Diagnosis supported by
  imaging findings (MRI, CT, bone scan) and clinical presentation.

**Plan**

- Surgical resection of the osteoid osteoma via a mini-open posterior approach at
  C7-T2 levels, using 2D/3D imaging guidance with Surgivisio® and surgical navigation.
- General anesthesia with prone positioning.
- Radiofrequency ablation not recommended due to proximity to dura mater.
- Vertebral stability to be preserved; no screw fixation required.
- Patient education provided regarding procedure details and expectations.
\end{verbatim}
\end{footnotesize}

\section{Prompts}
\label{app:prompts}

All prompts below are shared verbatim by both generator conditions and by both conditions' evaluator
calls; only the section-header renderer differs between backends (\S\ref{sec:scoped}). Long lines are
re-wrapped for the page; content is otherwise unmodified.

\paragraph{Generator, fresh generation (round 0).}
\begin{footnotesize}
\begin{verbatim}
You are an expert clinical medical scribe. Your task is to generate a structured SOAP
  note from a doctor-patient dialogue transcript.

STRICT 4-SECTION FORMAT:
Your output MUST contain exactly these 4 sections in order:
  **Subjective**
  **Objective**
  **Assessment**
  **Plan**

All 4 sections must be present in every note, even if data is limited.
Do NOT create additional sections.

SECTION GUIDELINES:

**Subjective** — Information reported by or about the patient:
- Chief Complaint: The patient's primary reason for the visit, in their own words
  (use "Not documented" if absent)
- History of Present Illness (HPI): Synthesise the patient's narrative about
  symptoms, onset, duration, severity, alleviating/aggravating factors, and
  relevant history mentioned in the dialogue
- Past Medical History: Conditions mentioned in the dialogue
- Medications: List medications discussed; note NEW vs CONTINUING where evident
- Allergies: Any mentioned
- Social/Family History: If discussed
- Review of Systems: Positives and relevant negatives mentioned in dialogue

**Objective** — Clinically documented findings from the encounter:
- Vital Signs: Values explicitly stated in the dialogue (do NOT infer)
- Physical Examination: Findings described by the clinician
- Diagnostic Results: Lab values, imaging findings, pathology mentioned
- Use ONLY information explicitly stated in the dialogue
- Do NOT include clinician impressions or interpretations here — only documented
  findings and test results
- Test result conclusions ARE documented findings and belong here, not in Assessment
  (e.g. "biopsy confirmed X", "MRI showed stenosis at C4/C5", "culture identified X")
- Do NOT use clinician qualifying language (likely, possibly, appears, consistent
  with)
  for your own inferences; if data is absent, omit

**Assessment** — Clinical interpretation and diagnoses:
- Primary diagnosis and supporting rationale from the dialogue
- Secondary diagnoses if mentioned
- Differential diagnoses if the clinician discusses them
- Synthesise findings from Subjective and Objective to support the diagnoses

**Plan** — Management decisions discussed:
- Diagnostic workup ordered (labs, imaging, referrals)
- Treatments initiated or changed (medications, procedures, interventions)
- Patient education discussed
- Follow-up instructions

DATA FIDELITY RULES:
- Include ONLY information explicitly present in the dialogue
- NEVER fabricate details, infer absent information, or fill narrative gaps
- If a section has minimal data, state it briefly rather than inventing content
- Use natural clinical prose — no raw codes, no templates, no filler phrases
- Preserve exact quantitative values (doses, measurements, dates, procedure names)
  verbatim as
  stated in the dialogue — do not paraphrase, round, or approximate them

STATUS AND TEMPORAL ACCURACY:
- State the completion status of every action exactly as the dialogue indicates, not
  as a generic
  SOAP convention would expect. If the dialogue shows a test, evaluation, or referral
    has ALREADY
  happened, describe it as completed (past tense) — do NOT describe it as "pending",
    "planned",
  "to be scheduled", or "will be arranged"
- Conversely, do not describe a future or ordered action as if it were already
  completed
- When uncertain about status, match the dialogue's own tense rather than defaulting
  to a
  conventional SOAP phrase

PLAN-SECTION FABRICATION GUARD:
- Do not add standard SOAP boilerplate to Plan (e.g. "follow-up in N weeks", "referral
  to
  [specialist] if needed", "patient education on [topic] provided", "continue to
    monitor") unless
  the dialogue explicitly discusses that specific action
- Generic management language that sounds clinically plausible but was not actually
  discussed is a
  fabrication, not a reasonable inference — every Plan item must be traceable to a
    specific
  statement in the dialogue
\end{verbatim}
\end{footnotesize}

\paragraph{Generator, section-scoped refinement (rounds $\geq 1$).}
\begin{footnotesize}
\begin{verbatim}
You are an expert clinical medical scribe reviewing and correcting a SOAP note that
  was generated from a doctor-patient dialogue.

You will be given:
1. The ORIGINAL DIALOGUE (source of truth)
2. Your PREVIOUS SOAP NOTE (the draft to refine)
3. EVALUATION FEEDBACK (specific issues to fix, prioritised by severity)
4. POSITIVE ANCHORS (sections/dimensions that already passed — preserve these)

YOUR TASK:
- Fix ONLY the specific issues listed in the feedback
- Preserve sections identified in positive_anchors — do NOT modify passing content
- Keep the same strict 4-section format: Subjective, Objective, Assessment, Plan
- All fixes must be grounded in the dialogue — do NOT invent information to fix errors
- If a claim cannot be fixed without fabricating data, remove it entirely

SECTION BOUNDARY DISCIPLINE (this applies even when the feedback is only about
missing facts, not about structure — do not let fixing one break the other):
- Every fact belongs in EXACTLY ONE section — the correct one for its type
  (Subjective: patient-reported information; Objective: documented data and
  findings only; Assessment: diagnoses and clinical interpretation only;
  Plan: actions and orders only).
- NEVER repeat or restate the same sentence, finding, or action in more than
  one section. When you add a fact to satisfy feedback, add it ONLY to the
  one section it belongs in.
- Assessment must contain ONLY diagnoses and interpretation — no treatment
  actions, orders, medications, or follow-up instructions. Those belong in
  Plan, and nowhere else.
- When adding a fact to satisfy feedback, first decide WHEN it happened:
  * Already performed / already known (past surgeries, prior treatments,
    test results, pathology/imaging findings, history) → Subjective (if
    patient-reported history) or Objective (if a documented result/finding).
    NEVER put an already-completed procedure, test result, or finding in Plan.
  * An action being ordered or started as a result of THIS encounter
    (new medication, new referral, new test to be done, follow-up) → Plan
    only.
- Before returning your answer, check each section against this rule and
  remove any sentence that duplicates content already present in, or that
  more properly belongs in, a different section.

STATUS AND TEMPORAL ACCURACY (a common source of factual-accuracy feedback):
- Check the completion status of every action against the dialogue exactly, not
  against SOAP
  convention. If the dialogue shows something has ALREADY happened, it must be
    described as
  completed — remove words like "pending", "planned", "to be scheduled", or "will be
    arranged" if
  the dialogue contradicts them
- Preserve exact quantitative values (doses, measurements, dates, procedure names)
  verbatim as
  stated in the dialogue rather than paraphrasing them

PLAN-SECTION FABRICATION GUARD (a common source of factual-accuracy feedback):
- If the feedback flags a Plan item as unsupported, do not rewrite it to sound more
  plausible —
  remove it unless you can point to the specific dialogue statement that supports it
- Do not introduce standard SOAP boilerplate ("follow-up in N weeks", "referral if
  needed",
  "patient education provided") to fill a section that the feedback did not ask you to
    expand

STRICT RULES:
- Return only the complete revised SOAP note — no commentary or preamble
- Maintain clinical prose quality throughout
- Do NOT add sections beyond the 4 required
\end{verbatim}
\end{footnotesize}

\paragraph{Factual verifier, claim extraction.}
\begin{footnotesize}
\begin{verbatim}
You are a clinical message analysis assistant. Your task is to identify the key
  factual claims in the provided SOAP note section that should be traceable back to
  what was said in a doctor-patient dialogue.

Instructions:
- A factual claim is any statement about the patient's symptoms, history, test
  results, diagnosis, treatment, medications, vital signs, or other clinical details.
- Include comparative statements (e.g. "remains stable", "consistent with prior
  visit") and any temporal references.
- Include absence/negation statements (e.g. "No drug allergies", "No prior surgeries")
  — these must be checked against the dialogue.
- Exclude vague general medical statements that do not make specific assertions about
  THIS patient.
- Exclude boilerplate placeholders (e.g. "Not documented", "Data unavailable").
- Group closely related measurements into a SINGLE claim.
- Keep the total number of claims concise — aim for no more than 15 per section.
- For each claim, return it as a numbered list item.

Return your answer as a numbered list:
1. <claim text>
2. <claim text>
...
\end{verbatim}
\end{footnotesize}

\paragraph{Factual verifier, claim extraction (Plan variant).}
\begin{footnotesize}
\begin{verbatim}
You are a clinical message analysis assistant. Your task is to identify the key
  verifiable claims in a SOAP note PLAN section that should be traceable back to what
  was said in a doctor-patient dialogue.

A verifiable PLAN claim is a specific action explicitly discussed in the dialogue: a
  test ordered, a medication prescribed or continued, a referral raised, a procedure
  mentioned, or a follow-up timeline agreed upon.

Instructions:
- Include specific named tests, medications, referrals, procedures, or follow-up
  timelines that appear in the Plan.
- Exclude generic non-verifiable boilerplate: "monitor symptoms", "supportive care",
  "educate the patient", "discuss findings", "keep patient informed" — these cannot
  be verified against the dialogue.
- Exclude forward-looking rationale or explanations (those belong in Assessment).
- Group closely related items into a SINGLE claim.
- Aim for no more than 10 claims.

Return your answer as a numbered list:
1. <claim text>
2. <claim text>
...
\end{verbatim}
\end{footnotesize}

\paragraph{Factual verifier, entailment check.}
\begin{footnotesize}
\begin{verbatim}
You are a clinical fact-checking assistant. Your task is to verify whether factual
  claims made in a SOAP note are strictly grounded in the content of a doctor-patient
  dialogue transcript.

Instructions:
- A claim is SUPPORTED (1) ONLY if:
  * The exact or equivalent information is explicitly stated in the dialogue, OR
  * The claim uses a different but clinically equivalent term for something directly
    stated (e.g. "elevated BP" when the dialogue reports a specific high blood
      pressure
    reading; "tachycardia" when the dialogue states "heart rate of 120"), OR
  * For absence/negation claims: the patient or clinician explicitly denies it in
    the dialogue (e.g. "denies allergies", "no prior surgeries mentioned")
  * Exception — boilerplate documentation placeholders such as "Not documented",
    "Not reported", "Not available" are always SUPPORTED; they record a
    documentation gap, not a clinical assertion about the patient
- A claim is NOT SUPPORTED (0) if:
  * It asserts a specific fact, value, or finding that is not explicitly stated or
    directly entailed by the dialogue — even if it is clinically plausible
  * It introduces a diagnosis, medication, or procedure not mentioned in the dialogue
  * It states an absence when the dialogue explicitly mentions the contrary
  * It is a clinical inference that goes beyond what the dialogue directly supports
    (e.g. inferring "hypertension" solely because antihypertensives were mentioned,
    without the dialogue naming or confirming the diagnosis)
- Do NOT assume SUPPORTED because something is clinically reasonable or typical.
  The standard is: "Is this explicitly stated or directly entailed by the transcript?"

Return ONLY valid JSON, no preamble:
{
  "claim_1": {"support": 1, "explanation": "..."},
  "claim_2": {"support": 0, "explanation": "..."}
}
\end{verbatim}
\end{footnotesize}

\paragraph{Completeness evaluator.}
\begin{footnotesize}
\begin{verbatim}
You are a clinical quality analyst reviewing whether a generated SOAP note adequately
  covers the clinical content of a doctor-patient dialogue.

Your task has two steps:

STEP 1 — Extract key clinical facts from the dialogue.
Identify ALL clinically important facts mentioned in the dialogue, grouped into these
  categories:
  - symptoms: Patient-reported symptoms, complaints, duration
  - findings: Physical exam findings, investigation results, imaging, labs
  - diagnoses: Diagnoses stated or strongly implied by the clinician
  - medications: Drugs mentioned (prescribed, changed, discontinued, current)
  - procedures: Any procedures performed, ordered, or discussed
  - vitals: Any vital signs values mentioned

STEP 2 — Check coverage in the SOAP note.
For each extracted fact, determine whether it is present in the SOAP note. Use
  semantic matching — accept synonyms and clinically equivalent expressions.

A fact is PRESENT if the SOAP note:
  * States it explicitly, or
  * Uses an equivalent clinical term (e.g. "hypertension" for "high blood pressure")

A fact is MISSING if it is absent from the SOAP note. In particular:
  * Individual lab values (e.g. Hgb 11.6, Na 133, ESR 33) must be explicitly named —
    a generic phrase such as "labs reviewed" or "elevated inflammatory markers" does
    NOT count as coverage for specific values mentioned in the dialogue.
  * Individual vital sign values (e.g. BP 145/90, HR 112) must appear explicitly.
  * Procedures and medications must be named, not implied by a general category.
  * Do NOT mark a fact as PRESENT simply because the SOAP note is clinically plausible
    or because the general topic is mentioned — the specific fact must be there.

Return ONLY valid JSON in this exact format:
{
  "extracted_facts": {
    "symptoms":   [{"fact": "...", "present": true|false}, ...],
    "findings":   [{"fact": "...", "present": true|false}, ...],
    "diagnoses":  [{"fact": "...", "present": true|false}, ...],
    "medications":[{"fact": "...", "present": true|false}, ...],
    "procedures": [{"fact": "...", "present": true|false}, ...],
    "vitals":     [{"fact": "...", "present": true|false}, ...]
  }
}
If a category has no facts in the dialogue, return an empty list for it.
\end{verbatim}
\end{footnotesize}

\paragraph{Structural fidelity evaluator (one chain shown).}
\begin{footnotesize}
\begin{verbatim}
You are a strict clinical documentation quality evaluator.
You will be given two sections of a SOAP note and a scoring rubric.
You MUST respond with valid JSON only — no markdown, no prose outside the JSON.

Response format:
{
  "score": <integer 0-3>,
  "explanation": "<one or two sentences citing specific evidence from the sections>"
}
\end{verbatim}
\end{footnotesize}

\paragraph{Hallucination categorizer.}
\begin{footnotesize}
\begin{verbatim}
You are a clinical data quality analyst reviewing whether a claim in a generated SOAP
  note is accurate relative to a doctor-patient dialogue transcript.

You will be given:
1. A CLAIM extracted from a generated SOAP note
2. The DIALOGUE TRANSCRIPT used as the source
3. A NOTE explaining why the claim could not be confirmed

Your task: Classify the accuracy issue into ONE category:

**INACCURATE_DATA**: The claim states something objectively wrong compared to the
  dialogue:
- CODE_CONFUSION: A clinical concept was misidentified
- WRONG_VALUE: Correct concept but incorrect numeric value or unit
- INVENTED_FINDING: References a finding or event not mentioned in the dialogue
- CONTRADICTS_TRANSCRIPT: Directly contradicts what was said

**MISSING_FROM_TRANSCRIPT**: The claim may reflect clinical reality but was not
  discussed in the dialogue:
- NOT_IN_DIALOGUE: Information not mentioned in the conversation
- IMPLICIT_ASSUMPTION: Clinically reasonable but never stated

**CLINICAL_REASONING**: A reasonable clinical conclusion from the dialogue:
- STANDARD_INFERENCE: Normal clinical summary (e.g., inferring a diagnosis)
- TEMPORAL_CONTEXT: Logical conclusion about disease progression
- IMPLICIT_CONTEXT: Information implied but not explicitly stated

**NEEDS_REVIEW**: Insufficient information to classify confidently

Respond with ONLY valid JSON, no other text:
{
  "category":
    "INACCURATE_DATA|MISSING_FROM_TRANSCRIPT|CLINICAL_REASONING|NEEDS_REVIEW",
  "subcategory": "...",
  "confidence": "HIGH|MEDIUM|LOW",
  "evidence": "specific dialogue content supporting your conclusion",
  "severity": "CRITICAL|MODERATE|MINOR",
  "reasoning": "brief explanation"
}
\end{verbatim}
\end{footnotesize}
\section*{NeurIPS Paper Checklist}

\begin{enumerate}

\item {\bf Claims}
    \item[] Question: Do the main claims made in the abstract and introduction accurately reflect the paper's contributions and scope?
    \item[] Answer: \answerYes{}
    \item[] Justification: The abstract and Introduction state the measured quantities themselves (a $0.008$ factual-entailment gap, $99.6\%$ vs.\ $98.4\%$ routed-system acceptance over $250$ encounters, and the $26.8\%\to83.2\%$ ablation contrast) rather than a qualitative summary, and the scope limits --- a synthetic corpus, one small model, one frontier baseline --- are named in the same paragraphs and elaborated in the Limitations and Future Work section.

\item {\bf Limitations}
    \item[] Question: Does the paper discuss the limitations of the work performed by the authors?
    \item[] Answer: \answerYes{}
    \item[] Justification: The Limitations and Future Work section discusses the synthetic-corpus setting, generalization to other small models and less structured tasks, and states explicitly that the acceptance gate certifies conformance to the paper's own rubric-based verifier rather than independently validated clinical correctness.

\item {\bf Theory assumptions and proofs}
    \item[] Question: For each theoretical result, does the paper provide the full set of assumptions and a complete (and correct) proof?
    \item[] Answer: \answerNA{}
    \item[] Justification: The paper contains no theorems or proofs. The two displayed formulas --- the acceptance predicate in the Method section and the section-scoped splice operator defining refinement --- are definitions of the method, not derived results requiring proof.

    \item {\bf Experimental result reproducibility}
    \item[] Question: Does the paper fully disclose all the information needed to reproduce the main experimental results of the paper to the extent that it affects the main claims and/or conclusions of the paper (regardless of whether the code and data are provided or not)?
    \item[] Answer: \answerYes{}
    \item[] Justification: Dataset composition and splits, the acceptance predicate and every evaluator's rubric and scoring formula (Appendix on evaluator rubrics), the full QLoRA fine-tuning configuration (Appendix on fine-tuning configuration), and every generator and evaluator prompt verbatim (Appendix on prompts) are given. 

\item {\bf Open access to data and code}
    \item[] Question: Does the paper provide open access to the data and code, with sufficient instructions to faithfully reproduce the main experimental results, as described in supplemental material?
    \item[] Answer: \answerNo{}
    \item[] Justification: No code repository, fine-tuned adapter, or data release accompanies this submission. The training and evaluation corpus is an existing public dataset (cited in the Experimental Setup section), but no artifact produced by this paper is released with an accompanying URL.

\item {\bf Experimental setting/details}
    \item[] Question: Does the paper specify all the training and test details (e.g., data splits, hyperparameters, how they were chosen, type of optimizer) necessary to understand the results?
    \item[] Answer: \answerYes{}
    \item[] Justification: Data splits, generator temperature, all four acceptance thresholds, the round budget, and the full QLoRA hyperparameters (rank, $\alpha$, dropout, quantization, optimizer, schedule, epochs) are given in the Experimental Setup section and the fine-tuning appendix.

\item {\bf Experiment statistical significance}
    \item[] Question: Does the paper report error bars suitably and correctly defined or other appropriate information about the statistical significance of the experiments?
    \item[] Answer: \answerYes{}
    \item[] Justification: Paired significance tests accompany the comparative claims central to the paper's argument: a Wilcoxon signed-rank test for paired factual-accuracy scores, an exact McNemar test for the routed-system-vs-frontier-only acceptance comparison, and a one-sided Mann--Whitney $U$ test for the kept-versus-escalated separation, each reported with its $p$-value and sample size in the Results section and the corresponding appendix.

\item {\bf Experiments compute resources}
    \item[] Question: For each experiment, does the paper provide sufficient information on the computer resources (type of compute workers, memory, time of execution) needed to reproduce the experiments?
    \item[] Answer: \answerNo{}
    \item[] Justification: On-device generation hardware is specified (Apple Silicon, $64$\,GB unified memory, no discrete GPU) and API-based verification cost is accounted for in full (call counts, token volumes, and dollar cost, in the end-to-end cost accounting appendix), but the compute type and wall-clock duration used to fine-tune the QLoRA adapter are not reported.

\item {\bf Code of ethics}
    \item[] Question: Does the research conducted in the paper conform, in every respect, with the NeurIPS Code of Ethics \url{https://neurips.cc/public/EthicsGuidelines}?
    \item[] Answer: \answerYes{}
    \item[] Justification: The paper uses a synthetic dialogue corpus containing no real patient data, involves no human subjects or crowdsourcing, and the submission is anonymized.

\item {\bf Broader impacts}
    \item[] Question: Does the paper discuss both potential positive societal impacts and negative societal impacts of the work performed?
    \item[] Answer: \answerNo{}
    \item[] Justification: The paper does not include a dedicated discussion of societal impact. The Limitations and Future Work section addresses methodological scope --- corpus realism, generalization, and the validity of the verifier's own certification --- rather than the positive or negative societal consequences of deploying such a system.

\item {\bf Safeguards}
    \item[] Question: Does the paper describe safeguards that have been put in place for responsible release of data or models that have a high risk for misuse (e.g., pre-trained language models, image generators, or scraped datasets)?
    \item[] Answer: \answerNA{}
    \item[] Justification: No model, adapter, or dataset is released as part of this submission, so no release safeguards apply.

\item {\bf Licenses for existing assets}
    \item[] Question: Are the creators or original owners of assets (e.g., code, data, models), used in the paper, properly credited and are the license and terms of use explicitly mentioned and properly respected?
    \item[] Answer: \answerNo{}
    \item[] Justification: The dialogue corpus and the base small-language model are cited by name and source, but their specific license terms are not stated in the paper text.

\item {\bf New assets}
    \item[] Question: Are new assets introduced in the paper well documented and is the documentation provided alongside the assets?
    \item[] Answer: \answerNA{}
    \item[] Justification: No new dataset, code, or model checkpoint is released alongside this paper.

\item {\bf Crowdsourcing and research with human subjects}
    \item[] Question: For crowdsourcing experiments and research with human subjects, does the paper include the full text of instructions given to participants and screenshots, if applicable, as well as details about compensation (if any)?
    \item[] Answer: \answerNA{}
    \item[] Justification: This paper conducts no crowdsourcing and involves no human subjects; the underlying dialogue corpus is an existing, previously released dataset used as-is.

\item {\bf Institutional review board (IRB) approvals or equivalent for research with human subjects}
    \item[] Question: Does the paper describe potential risks incurred by study participants, whether such risks were disclosed to the subjects, and whether Institutional Review Board (IRB) approvals (or an equivalent approval/review based on the requirements of your country or institution) were obtained?
    \item[] Answer: \answerNA{}
    \item[] Justification: No human-subjects research was conducted by the authors; the paper uses an existing synthetic corpus rather than collecting new data from participants.

\item {\bf Declaration of LLM usage}
    \item[] Question: Does the paper describe the usage of LLMs if it is an important, original, or non-standard component of the core methods in this research? Note that if the LLM is used only for writing, editing, or formatting purposes and does \emph{not} impact the core methodology, scientific rigor, or originality of the research, declaration is not required.
    \item[] Answer: \answerYes{}
    \item[] Justification: Large language models are not an incidental writing aid but the core object of study: both candidate generators (GPT-4.1 and a fine-tuned Gemma-3-4B) and every evaluator in the verification stack are LLMs, fully specified in the Method and Experimental Setup sections and the evaluator-rubric and prompts appendices.

\end{enumerate}

\end{document}